\documentclass[conference]{IEEEtran}
\IEEEoverridecommandlockouts

\usepackage{cite}
\usepackage{hyperref}
\hypersetup{
    colorlinks=true,
    linkcolor=black,
    filecolor=black,      
    urlcolor=black,
    citecolor=black,
    }
\usepackage{amsmath,amssymb,amsfonts}
\usepackage{cleveref}

\usepackage{algorithm}
\usepackage{algpseudocode}
\usepackage{graphicx}
\usepackage{textcomp}
\usepackage{xcolor}
\usepackage{tikz}
\usepackage{tikz-3dplot}
\usetikzlibrary{arrows.meta, calc}
\usepackage{subfig}
\usepackage{caption}
\usepackage{cleveref}
\usepackage{booktabs}
\def\BibTeX{{\rm B\kern-.05em{\sc i\kern-.025em b}\kern-.08em
    T\kern-.1667em\lower.7ex\hbox{E}\kern-.125emX}}

\newcommand{\reffig}[1]{Fig.~\ref{#1}}

\newcommand{\refsec}[1]{Sec.~\ref{#1}}
\newcommand{\reftab}[1]{Table~\ref{#1}}

\algrenewcommand\algorithmicrequire{\textbf{Input:}}
\algrenewcommand\algorithmicensure{\textbf{Output:}}

\begin{document}

    \title{Autonomous Inspection of Power Line Insulators with UAV on an Unmapped Transmission Tower\\
\thanks{This work was funded by the Czech Science Foundation (GAČR) under research project no. 26-22606S and by European Union under the project Robotics and advanced industrial production (reg. no. CZ.02.01.01/00/22\_008/0004590).}
}

\author{\IEEEauthorblockN{Václav Riss, Vít Krátký, Robert Pěnička, Martin Saska}
}

\maketitle
\begin{abstract}

This paper introduces an online inspection algorithm that enables an autonomous UAV to fly around a transmission tower and obtain detailed inspection images without a prior map of the tower.
Our algorithm relies on camera-LiDAR sensor fusion for online detection and localization of insulators.
In particular, the algorithm is based on insulator detection using a convolutional neural network, projection of LiDAR points onto the image, and filtering them using the bounding boxes.
The detection pipeline is coupled with several proposed insulator localization methods based on DBSCAN, RANSAC, and PCA algorithms. 
The performance of the proposed online inspection algorithm and camera-LiDAR sensor fusion pipeline is demonstrated through simulation and real-world flights.
In simulation, we showed that our single-flight inspection strategy can save up to 24 \% of total inspection time, compared to the two-flight strategy of scanning the tower and afterwards visiting the inspection waypoints in the optimal way.
In a real-world experiment, the best performing proposed method achieves a mean horizontal and vertical localization error for the insulator of 0.16 $\pm$ 0.08 m and 0.16 $\pm$ 0.11 m, respectively. 
Compared to the most relevant approach, the proposed method achieves more than an order of magnitude lower variance in horizontal insulator localization error.
\end{abstract}
{\footnotesize
\noindent \textbf{ } 
\noindent \textbf{Verification Video:} \url{https://youtu.be/n7C9efFiVeo}\\
\vspace{-0.7em}
}


\section{Introduction}


Every high-voltage power line worldwide must be regularly inspected, which involves inspecting the transmission towers, conductors, and other essential components.
One of the most important components to be inspected is insulators, which can become contaminated or physically damaged due to long-term exposure to outdoor conditions, increasing the risk of complete mechanical or electrical failure. 
To prevent potential adverse impacts associated with power outages, regular inspection of insulators is required.
Moreover, inspecting these components is much cheaper than the potential financial loss and cost of repairs associated with power outages. 

Inspection of insulators is difficult due to factors such as the height, the absence of ground access roads, and the variety of tower types with different shapes and dimensions.
Nowadays, the task is carried out through labor-intensive on-foot inspections with binoculars or performed by manually controlled Unmanned Aerial Vehicle (UAV) flying along power lines to capture inspection images.
With the arrival of newer technologies such as high-capacity batteries, higher computational power, which allows deployment of computer vision algorithms on smaller computers, and ongoing research in the field of autonomous UAVs, new ways of inspecting insulators are being developed.
The existing methods use UAVs running trained neural networks to detect insulators in the obtained images \cite{8853298, renwei2021key} in order to estimate the condition of the insulator and measure the GNSS location \cite{rs13020230}. In \cite{8885596}, the authors introduce an UAV system that allows autonomous power line and transmission tower inspection.
Another work \cite{rs15030865} introduces an UAV system capable of autonomous recharging, image acquisition, and detection of bird nests. Although current systems enable autonomous insulator inspection, they still require prior knowledge of the inspected tower structure.
\begin{figure}[t]
  \centering
  \includegraphics[width=0.49\textwidth]{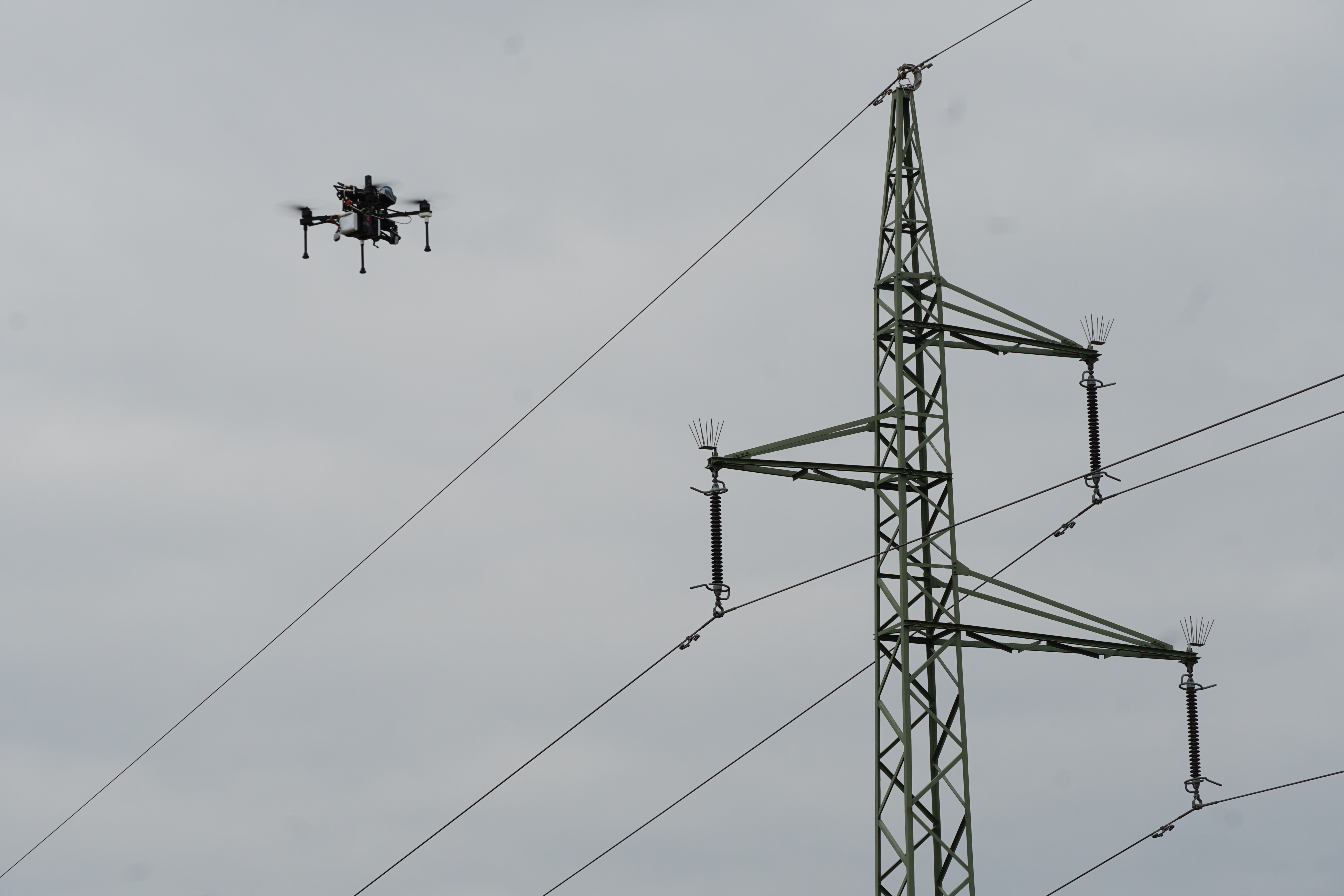}  
  \caption{Deployment of the proposed pipeline for autonomous inspection of insulators onboard a UAV in a real-world scenario.}
  \vspace{-2em}
  \label{fig:realworld_intro}
\end{figure}
\vspace{-1em}

This paper presents an online inspection algorithm for the inspection of unmapped transmission towers within a single flight without any information about the exact position or number of insulators.
Onboard insulator detection and its localization pipeline based on camera-LiDAR sensor fusion enables the UAV to detect and localize insulators attached to the transmission tower and acquire inspection images from optimal inspection waypoints.
Deployment of the pipeline onboard a UAV in an online fashion enables performing the inspection in one flight without initial mapping, being more time efficient than a two-flight approach --- first scanning the tower to create a map for finding the insulator locations, and subsequently flying along an optimal precomputed trajectory to acquire images of the insulators.
Our method is based on insulator detection using a convolutional neural network YOLOv11n, projection of LiDAR points onto the image, and filtering them using the bounding boxes. 
For insulator localization, several different methods based on DBSCAN \cite{ester1996density}, RANSAC \cite{fischler1981random}, and PCA \cite{pearson1901liii} are proposed.

We show the performance of the insulator detection and localization together with the online inspection algorithm in both simulation and real-world scenarios.
The online inspection algorithm is evaluated and compared to the two-flight approach in simulation in terms of the duration of the inspection mission. 
The performance of the proposed online inspection algorithm significantly depends on the number of inspection waypoints that need to be visited in order to capture inspection images.
For 8 inspection waypoints in sample scenarios, our method can save roughly 24 \% of the total inspection time and is still the more time efficient one for up to 20 inspection waypoints. 
The best performing insulator localization method, DBSCAN followed by RANSAC (DBSCAN+RANSAC), achieves an insulator localization error of $0.16 \pm 0.08$ m in the $xy$ plane and $0.16 \pm 0.11$ m in altitude $z$ in the real-world experiment. This shows that our approach yields sufficient precision of insulator localization for the acquisition of the inspection images. Moreover, our approach achieves over an order of magnitude lower insulator localization error variance than the approach based on a binocular camera \cite{rs13020230}.

\section{Related Work}


\subsection{Insulator Detection from Image Data}


Camera-based approaches are commonly used for insulator detection, but the task remains challenging due to significant visual variability \cite{zhang2010simple, wu2013active, 8853298, renwei2021key, jing2024insulator,10298815,10413457}. This difficulty arises from complex backgrounds in aerial images, including vegetation, rivers, roads, and buildings. Furthermore, the appearance of insulators varies substantially because of different insulator types, changing illumination conditions, and variations in camera viewpoints during inspection.

Earlier methods for detecting insulators from images mostly relied on traditional image processing techniques.
For example, a method in \cite{zhang2010simple} detects tempered glass insulators by converting aerial images to grayscale using the hue, saturation, intensity color model and applying a fixed threshold based on the mean intensity of connected components to obtain a binary image. 
Another method \cite{wu2013active} involves using an active contour model based on texture distribution to detect inhomogeneous insulators. While these techniques show promising results, they have some drawbacks, such as high computational cost, the need for manual parameter initialization, and poor adaptability to illumination changes and varying light conditions.

Recent research on insulator detection has shifted toward advanced deep learning detectors and instance segmentation models.
For example, a framework You Only Look Once (YOLO) \cite{redmon2016lookonceunifiedrealtime}, specifically the YOLOv2 model, is proposed in \cite{8853298} to detect insulators under the conditions of an uncluttered background.
In \cite{renwei2021key}, a modified version of YOLOv3 was specifically used to detect insulators.
The version has fewer convolutional layers, which increases speed without sacrificing accuracy, as those layers are not essential for detecting small objects.
Another notable direction is instance segmentation, where Mask R-CNN \cite{9585716} has been applied to insulator detection.
Recent papers use refined versions of modern YOLO versions for improved feature extraction \cite{jing2024insulator,10298815,10413457}. Methods introduced in \cite{jing2024insulator, 10413457} focus on missing sheds detection, in \cite{10298815} detector distinguishes between normal, self-explosive, damaged, and flashover insulators. 

Even though modern approaches demonstrate strong performance, their applicability is limited by the need to create dedicated datasets for training the models, especially instance segmentation models, which require precise pixel-level annotations. 
Those methods also cannot be utilized for precise insulator localization, as they would only rely on the insulator in the image and the predefined size of the insulator, which is an unrealistic assumption in the majority of target real-world power-line inspection scenarios.

\subsection{Detection and Localization of Insulators Using Binocular Camera}
Methods dealing with insulator detection and localization based on images from a binocular camera are rare in the literature.
A recent paper published in \cite{rs13020230} presents an automatic inspection system that detects and localizes insulators using a binocular camera setup. The system obtains spatial information about the insulator using depth information. The paper concludes that localization accuracy is sufficient for the autonomous insulator inspection task. However, the real-world results show the mean insulator localization errors are 0.92 $\pm$ 3.72 m for longitude, 0.79 $\pm$ 5.86 m for latitude, and 0.03 $\pm$ 0.29 m for altitude.

\subsection{Detection and Localization of Insulators from Point Cloud}

Methods detecting insulators from LiDAR data \cite{drones8060241, 10587504} first scan entire transmission towers and then identify the sections that correspond to insulators, and thus localize insulators. The method presented in \cite{drones8060241} detects insulators in point cloud data based on a set of histogram-derived features: horizontal density, horizontal void, horizontal width, vertical width, and vertical void. In \cite{10587504}, the authors suggest using the DBSCAN algorithm to first identify regions in the point cloud that contain towers. Then, features describing the shape and spatial distribution of the points are calculated and weighted using the entropy weighting method. These features are then used to train a random forest classifier, which adapts its parameters automatically. 

Although the described methods are able to precisely detect and locate insulators, their weakness is the inefficiency of potential capturing images of insulators mounted on the inspected tower. To obtain the detailed images of the insulators, the UAV has to fly around the tower at least two times: the first time to get the complete point cloud, and subsequently fly to capture the images. Our approach can complete the inspection in one flight by detecting and locating insulators online.


\subsection{Power Lines Inspection Tasks}

Existing works \cite{9476742,rs15030865, rs13020230, xing2023autonomous} introduce power line infrastructure detection as part of the UAV autonomous inspection systems. In \cite{9476742}, the authors describe a system that integrates multiple sensors and automates many tasks, such as detection, tracking, and identification of infrastructure, including insulators from a top-down view.
Another work \cite{rs15030865} focuses on the development of a rechargeable UAV system that detects bird nests and obtains images of infrastructure. Acquisition of insulator inspection images together with spatial information is also introduced in the UAV system presented in \cite{rs13020230}. Method in \cite{xing2023autonomous} proposes a model predictive controller that tightly couples perception and control for UAV-based power line inspection. In \cite{11072833}, authors present multiple diverse UAV platforms suitable for power line inspection tasks and show their deployment in large-scale mock-up scenarios.

The presented work shows great potential to save a considerable amount of time compared to manual inspection flights \cite{rs15030865}. However, there is still significant room for improvement.
Current solutions for insulator inspection work with precise information about transmission towers \cite{rs15030865, 8885596} and capture the inspection images from precomputed positions \cite{rs15030865}. 
Our approach can complete the inspection using only minimal prior information about the transmission tower (maximum height, width, and GPS coordinates), typically known to the grid owner, and compute a suitable position for onboard acquisition of insulator inspection images. This reduces the required amount of input data, minimizes the number of flights, and accelerates the data acquisition process.


\section{Methodology}

\begin{figure}[htbp]
  \centering
  \input{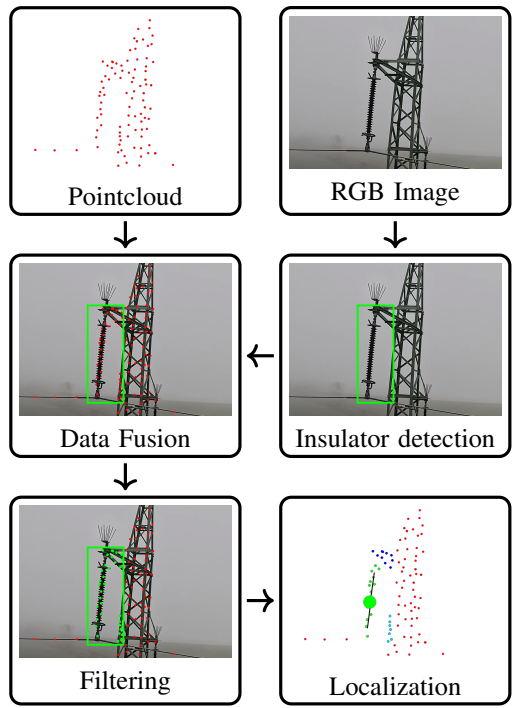}
  \caption{Illustration of the insulator detection and localization algorithmic pipeline. Points inside the bounding box marked in non-red color are clustered and utilized for insulator position and orientation estimation.}
  \label{fig:detectionlocalization_diagram}
  \vspace{-0.5cm}
\end{figure}

The proposed algorithm for insulator detection and localization by camera and LiDAR data fusion consists of the following steps. 
First, the RGB image is used to detect an insulator with a CNN-based detector. 
Next, the points are projected onto the image plane, and those within the detection bounding box are considered for later processing.
The position of the insulator is estimated by the clustering and line-fitting methods.
An illustration of the algorithm is depicted in \reffig{fig:detectionlocalization_diagram}.

The estimated position and orientation of the insulator are further used in our proposed online inspection algorithm for the acquisition of detailed inspection images.
The algorithm guides the UAV along an exploration path around the tower.
Whenever an insulator is detected and localized, the algorithm computes inspection waypoints for the acquisition of the inspection images.
 The UAV visits inspection waypoints and continues in flight around the tower until the flight is finished.


\subsection{Insulator Detection and Dataset Description}\label{sec:insulator_detection}

The YOLO object detection framework was selected for insulator detection due to its high detection accuracy, robustness and computational efficiency, which are important for real-time detection. In this work, the lightweight YOLOv11n variant \cite{yolo11_ultralytics} was used as the detection model, as it is the best-performing YOLO variant with low computational requirements. To further reduce inference time, the trained model was optimized using TensorRT, resulting in shorter inference time with only a negligible impact on accuracy.

\begin{figure}[htbp]
  \centering
  \subfloat[Tower A] {
    \includegraphics[width=0.21\textwidth]{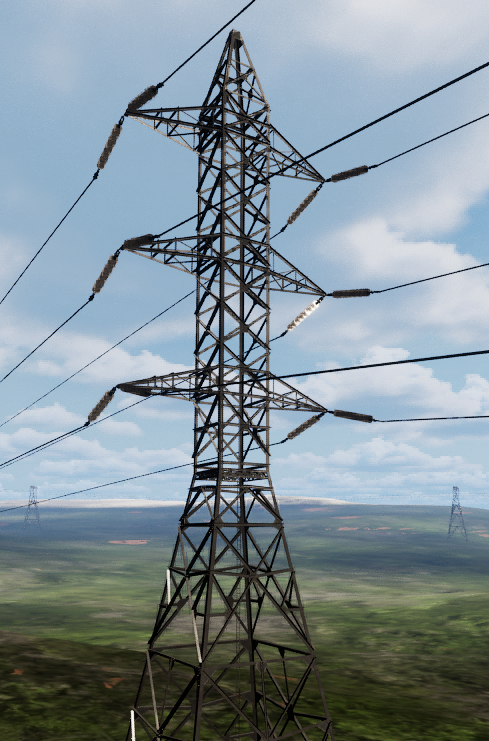}
    \label{fig:tower_a}
  }
  \subfloat[Tower B] {
    \includegraphics[width=0.20\textwidth]{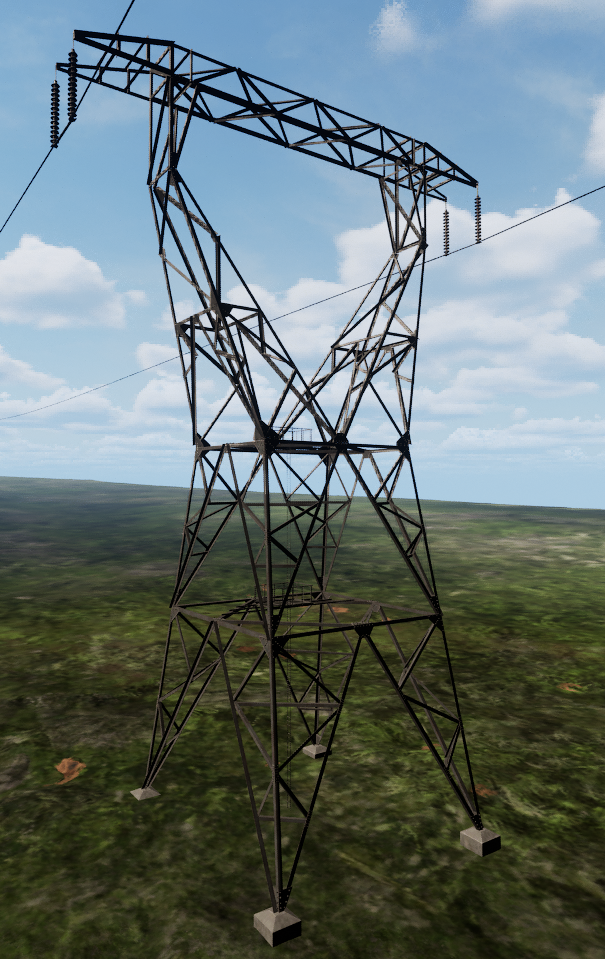}
    \label{fig:tower_b}
  }
  \caption{Simulated towers in Flight Forge simulator \cite{capek2025flightforge}, (a) shows tower A with twelve horizontal insulators, and (b) shows tower B with four vertical insulators.}
  \label{fig:towers}
  \vspace{-0.5cm}
\end{figure}

Since the shape of insulators and the simulation environment itself significantly differ from real-world insulators and conditions, two separate datasets were created for training, validation, and testing - one dataset for the simulation experiments, and one for the real-world deployment.
This approach results in two separate trained weight sets.
The simulation dataset was generated using the Flight Forge simulator  \cite{capek2025flightforge}.
Specifically, the environment was composed of towers A with twelve nearly horizontal insulators and type B towers with four vertical insulators (see \reffig{fig:towers}), which were used for data collection and testing.
The simulation dataset consists of approximately 1500 images and is split into training, test, and validation sets in a ratio of 0.8 $\mathrel{:}$ 0.1 $\mathrel{:}$ 0.1.  
To increase dataset variability, simulator settings such as time of day and weather conditions (sunny, cloudy, and rainy) were varied. 
The real-world training and validation datasets were composed of approximately 3200 images collected during multiple flights and randomly split in 0.85 $\mathrel{:}$ 0.15 ratio.
Training data were augmented by changing contrast, brightness, and rotation.
The test dataset consisted of images acquired on a different day.
Both the simulation and real-world datasets were trained for 200 epochs and evaluated using the Ultralytics framework \cite{yolo11_ultralytics}.

\vspace{-0.1cm}

\subsection{Fusion of Camera and Point Cloud Data}

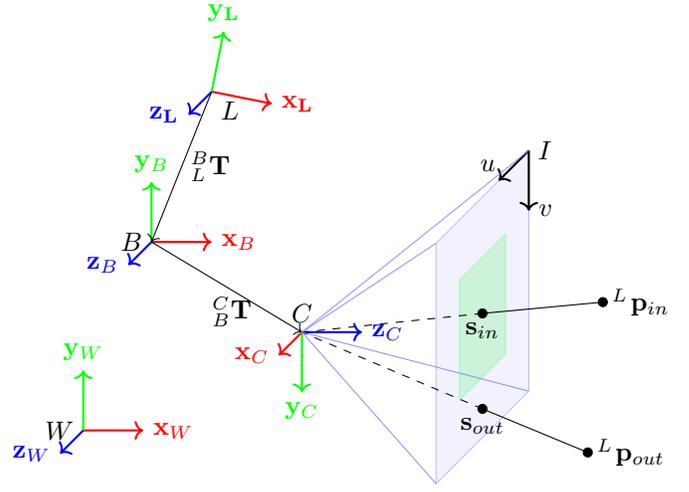
\begin{figure}[htbp]
  \centering
  \begin{tikzpicture}[scale=2,
    axis/.style={->, thick},
    frame/.style={axis},
    transform/.style={thick},
    ray/.style={thick, dashed},
]

\coordinate (W) at (0,0,0);

\draw[frame, red]   (W) -- ++(0.4,0,0) node[right] {$\mathbf{x}_W$};
\draw[frame, green] (W) -- ++(0,0.4,0) node[above] {$\mathbf{y}_W$};
\draw[frame, blue]  (W) -- ++(0,0,0.4) node[left]  {$\mathbf{z}_W$};

\node[left] at (W) {$W$};

\coordinate (B) at (0.8,1.6,0.9);


\begin{scope}[shift={(B)}]
    \draw[frame, red]   (0,0,0) -- ++(0.4,0,0) node[right] {$\mathbf{x}_B$};
    \draw[frame, green] (0,0,0) -- ++(0,0.4,0) node[above] {$\mathbf{y}_B$};
    \draw[frame, blue]  (0,0,0) -- ++(0,0,0.4) node[left] {$\mathbf{z}_B$};
\end{scope}

\node[left] at (B) {$B$};

\coordinate (C) at (1.8,1.0,0.9);

\draw[->, transform, thin] (B) -- (C) node[midway,below, xshift=2pt] {${}^{C}_{B}\mathbf{T}$};
\begin{scope}[shift={(C)}]
    \draw[frame, red]   (0,0,0) -- ++(0,0,0.4) node[left] {$\mathbf{x}_C$};
    \draw[frame, green] (0,0,0) -- ++(0,-0.4,0) node[below] {$\mathbf{y}_C$};
    \draw[frame, blue]  (0,0,0) -- ++(0.4,0,0) node[right] {$\mathbf{z}_C$};
\end{scope}

\node[above] at (C) {$C$};

\begin{scope}[shift={(0,0.5,0)}]
    \coordinate (I1) at (3.0,0.2,0.5);
    \coordinate (I2) at (3.0,1.0,0.5);
    \coordinate (I3) at (3.0,1.0,1.3);
    \coordinate (I4) at (3.0,0.2,1.3);
    
    \coordinate (J1) at (3.0,-0.2,0.1);
    \coordinate (J2) at (3.0, 1.4,0.1);
    \coordinate (J3) at (3.0, 1.4,1.7);
    \coordinate (J4) at (3.0,-0.2,1.7);
\end{scope}
\draw[fill=green!30, opacity=1.0,draw=green!50] (I1) -- (I2) -- (I3) -- (I4) -- cycle;
\draw[fill=blue!10, opacity=0.5, draw=blue!50] (J1) -- (J2) -- (J3) -- (J4) -- cycle;

\draw[->,thick] (J2) -- ++(0,-0.4,0) node[right] {$v$};
\draw[->,thick] (J2) -- ++(-0.2,-0.2,0) node[above, xshift=-4pt] {$u$};

\node[right] at (J2) {$I$};

\coordinate (p_in) at (3.8, 1.2,0.9);
\coordinate (p_out)  at (3.7, 0.2,0.9);

\coordinate (p_in_proj) at (3.0, 1.125,0.9);
\coordinate (p_out_proj)  at (3.0, 0.49,0.9);

\filldraw[black] (p_in) circle (0.03) node[right] {$^{L\textbf{ }}\!\mathbf{p}_{in}$};
\filldraw[black] (p_out) circle (0.03) node[right] {$^{L\textbf{ }}\!\mathbf{p}_{out}$};

\filldraw[black] (p_in_proj) circle (0.03) node[below] {$\mathbf{s}_{in}$};
\filldraw[black] (p_out_proj) circle (0.03) node[below] {$\mathbf{s}_{out}$};

\draw[thin] (p_in) -- (p_in_proj);
\draw[thin] (p_out) -- (p_out_proj);

\draw[ray, thin] (p_in_proj) -- (C);
\draw[ray, thin] (p_out_proj) -- (C);


\draw[thin, blue!50, opacity=0.85] (C) -- (J3);
\draw[thin, blue!50, opacity=0.85] (C) -- (J4);
\draw[thin, blue!50, opacity=0.85] (C) -- (J1);
\draw[thin, blue!50, opacity=0.85] (C) -- (J2);

\coordinate (L) at (1.2,2.6,0.9);

\draw[->, transform, thin] (L) -- (B) node[midway,right] {${}^{B}_{L}\mathbf{T}$};

\begin{scope}[shift={(L)}]
    \draw[frame, red]   (0,0,0) -- ++(0.4,-0.08,0) node[right] {$\mathbf{x_L}$};
    \draw[frame, green] (0,0,0) -- ++(0.08,0.4,0) node[above] {$\mathbf{y_L}$};
    \draw[frame, blue]  (0,0,0) -- ++(0,0,0.4) node[left] {$\mathbf{z_L}$};
\end{scope}

\node[below right] at (L) {$L$};

\end{tikzpicture}
  \caption{A scheme representing the LiDAR frame $L$, the body frame $B$, and the camera frame $C$. Fixed transformations between the frames allow projection of points ${}^{L}{\mathbf{p}}_{in}$ and ${}^{L}{\mathbf{p}}_{out}$ the image plane frame defined by I. Projected points ${\mathbf{s}}_{in}$ and ${\mathbf{s}}_{out}$ are then filtered out if they do not lie inside a green bounding box corresponding to detected insulator.}
  \label{fig:projection_and_filtration}\vspace{-2em}
\end{figure}

Fusion of LiDAR data and the RGB image data is an essential part of the perception pipeline, as the fused point cloud in the camera image can subsequently be filtered using detection bounding boxes to obtain spatial information. 
The fixed extrinsic transformations between the LiDAR frame $L$, the UAV body frame $B$, and the camera frame $C$ are used for the fusion.
Let ${}^{L}\tilde{\mathbf{p}}_{in}$ denote a point expressed in homogeneous coordinates in frame $L$.
Let ${}^{C}_{B}\mathbf{T}$ and ${}^{B}_{L}\mathbf{T}$ represent the known rigid-body transformation matrices from frame B to C, and from frame L to B, respectively.
The point is transformed into the camera frame $C$ using the well-known chain of homogeneous transformations
\begin{equation}
{}^{C}\tilde{\mathbf{p}}_{in}
=
{}^{C}_{B}\mathbf{T}\,
{}^{B}_{L}\mathbf{T}\,
{}^{L}\tilde{\mathbf{p}}_{in}.
\end{equation}
Point ${}^{C}\tilde{\mathbf{p}}_{in} =\left({}^{C}{p}_{in_x}, {}^{C}{p}_{in_y}, {}^{C}{p}_{in_z}, 1\right)$ is transformed into the image plane frame $I$ as 
\begin{equation}
u = f_x \frac{{}^{C}p_{in_x}}{{}^{C}p_{in_z}}, 
\qquad
v = f_y \frac{{}^{C}p_{in_y}}{{}^{C}p_{in_z}},
\end{equation}
where $f_x$ and $f_y$ are the focal lengths of the camera, and $u$, $v$ represent the coordinates of the projected point $s_{in} = (u, v)$ in the image frame $I$. The simulation uses a simplified camera model \cite{szeliski2022computer} with information about focal lengths $f_x$ and $f_y$ without lens distortion. For real-world applications, calibration is done in order to model radial and tangential distortion \cite{szeliski2022computer}. The projection is depicted in \reffig{fig:projection_and_filtration}.

\subsection{Point Cloud Filtering}

Filtering the point cloud is necessary for the insulator to be located, as the majority of original points do not correspond to reflections from the insulator itself.
For each detection, the 2D pixel coordinates within the bounding box produced by the insulator detection module are determined. 
Using the identified 2D pixel coordinates, the original 3D point cloud is filtered to retain only the points that project onto the bounding box region. 
This yields a subset of points that mostly correspond to the insulator. 
To obtain a denser LiDAR point cloud, three consecutive detection events along with their filtered point clouds are cumulated for the localization of the insulator.


\subsection{Localization of Insulators}\label{sec:localization_of_insulator}

We propose methods for the localization of insulators based on point cloud clustering and estimation of orientation using DBSCAN, RANSAC, and PCA. 
The methods take the filtered point cloud containing only points within the bounding box region and further process them to filter out remaining points corresponding to tower construction near the insulator. Next, they estimate the orientation and exact location of the insulator using an assumption on its cylindrical shape.

The filtered point cloud is first segmented using a clustering algorithm (DBSCAN or RANSAC).
This produces a set of spatially coherent clusters representing potential objects near the transmission line, including the insulator.
Among all clusters, the nearest one to the UAV is selected, and its center and orientation are calculated. 
The orientation of the insulator is estimated via PCA or RANSAC using the points of the selected cluster. 
The resulting line defines a direction that approximates the insulator’s longitudinal axis.
Subsequently, the center of the cluster is computed by using the line, which serves as a reference for extracting points that truly belong to the insulator. We assume that points located within a distance 
$\tau$ from the reference line correspond to the insulator. 
One of the proposed methods relying on PCA is designed to iteratively filter out possible bias caused by a considerable number of outliers. 
A new center is subsequently computed from this refined set of points, producing the final localized position of the insulator. 

In our work, we propose four different approaches for insulator localization. The first method is a simple DBSCAN-based clustering strategy. The second method relies solely on RANSAC for cluster and line estimation. The third approach combines DBSCAN with RANSAC (DBSCAN+RANSAC), as presented in Algorithm \ref{alg:dbscan_ransac}. Finally, the fourth method is a DBSCAN followed by PCA (DBSCAN+PCA) refinement procedure shown in Algorithm \ref{alg:dbscan_pca}.

\begin{algorithm}[htbp]

\footnotesize
\caption{DBSCAN + RANSAC}
\label{alg:dbscan_ransac}

\begin{algorithmic}[1]
\Require $point\_cloud$ in $B$ frame $, \;\tau$\Comment{$\tau$ = distance threshold}
\Ensure $\{center,\; orientation\}$ 

\Statex \hspace{-\algorithmicindent}\rule{\linewidth}{0.4pt}

\State $clusters$ $\gets$ DBSCAN($point\_cloud$)
\State $nearest$ $\gets \arg\min_{c \in clusters} \| \text{center}(
c) \|$
\State $(line,\; inliers) \gets$ RANSAC($nearest$)
\State $center$ $\gets$ median(projection($inliers$, $line$, $\tau$))
\State $orientation \gets $direction($line$)
\State \Return $\{center,\; orientation\}$
\end{algorithmic}
\end{algorithm}

\begin{algorithm}[htbp]
\footnotesize
\caption{DBSCAN + PCA}
\label{alg:dbscan_pca}

\begin{algorithmic}[1]
\Require $point\_cloud$ in $B$ frame $, \;\tau$ \Comment{$\tau$ = distance threshold}
\Ensure $\{center,\; orientation\}$
\Statex \hspace{-\algorithmicindent}\rule{\linewidth}{0.4pt}
\State $clusters \gets \text{DBSCAN}(point\_cloud)$
\State $nearest \gets \arg\min_{c\in clusters}\|\text{center}(c)\|$
\State \texttt{// Initial center $c$ and orientation $v$}
\State $v, c \gets \text{PCA}(nearest), \text{centroid}(nearest)$
\State $S \gets \{p\in nearest \mid \text{dist}(p,\text{line}(v,c))\le\tau\}$
\State $SC \gets \text{DBSCAN}(S)$
\If{$|\;SC\;|>1$} 
\State \texttt{// Reiterate on selected cluster $T$}
  \State $T \gets \text{largest}(SC)$
  \State $c',v' \gets \text{center}(T), \text{PCA}(T)$
  \State $S' \gets \{p\in nearest \mid \text{dist}(p,(c',v'))\le\tau\}$
  \State $center \gets \text{median}(\text{projection}(S',(c',v'),\tau))$
  \State $orientation \gets v'$
  \State \Return $\{center,\; orientation\}$
\Else
  \State $center \gets c$; $orientation\gets v$
  \State \Return $\{center,\; orientation\}$
\EndIf
\end{algorithmic}
\end{algorithm}
\subsection{Online Inspection Algorithm}\label{sec:inspection_planning_algorithm}
\begin{figure}[h]
  \centering
  \includegraphics[width=0.46\textwidth]{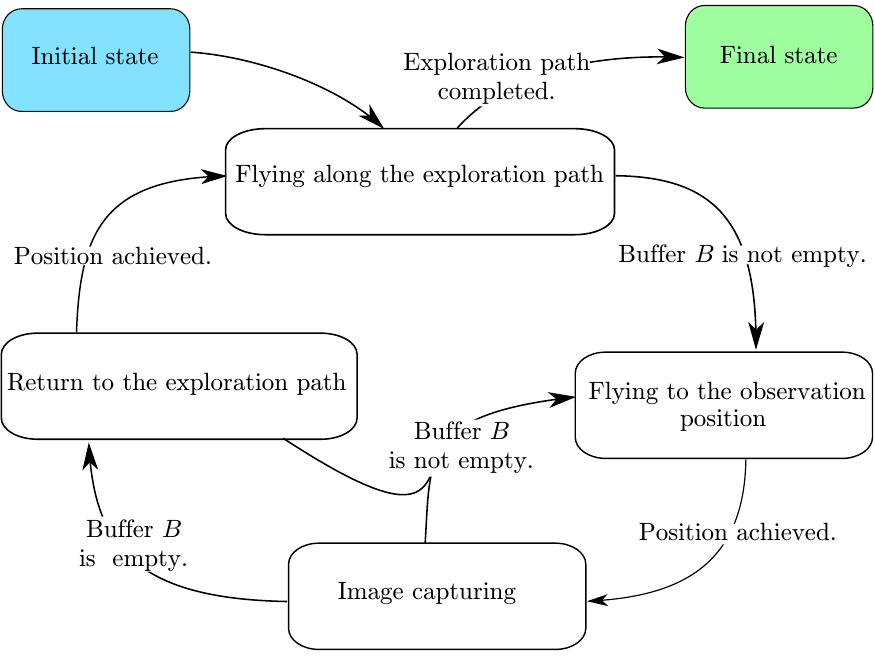}
  \caption{Illustrative scheme of the proposed state machine for inspection of insulators. Whenever an insulator is detected, inspection waypoints are saved into the buffer $B$. State machine empties $B$ by visiting the stored inspection waypoints. }
  \label{fig:state_machine}\vspace{-0.5cm}
\end{figure}

The insulator localization introduced in \refsec{sec:localization_of_insulator} is employed within the proposed online inspection algorithm.
The goal of the proposed algorithm is to autonomously acquire inspection images of insulators.
The scheme of the proposed state machine for autonomous inspection is depicted in \reffig{fig:state_machine}.

The UAV computes an exploration path around both sides of the tower from the approximate GPS coordinates of the tower being inspected and the coordinates of neighboring towers.
The coordinates of neighboring towers are used to estimate the orientation of the inspected tower and to define bounding volumes that help ensure safe flight planning, preventing the UAV from entering areas between towers where collisions with power lines could occur.
Subsequently, the UAV flies along the exploration path until an insulator is detected and localized. 
Whenever a new insulator is localized, the program computes inspection waypoints for capturing required inspection images from the extracted spatial information of the insulator.
Next, the UAV flies to the inspection waypoints and captures the required images of the insulator. Trajectories in our approach are generated by the polynomial trajectory planner introduced in \cite{richter2016polynomial}.
To avoid a collision with the tower, the maximum width and height of the tower are used to check whether the planned trajectory violates the predefined safety region around the tower.
After visiting all inspection waypoints, the UAV returns to the exploration path and continues the exploration.
The process ends when the UAV finishes the exploration path. 

Our approach needs only minimum structural information about the tower - specifically its maximum height, maximum width, and the GPS coordinates of the tower and of neighboring towers.
These parameters provide sufficient information for safe flight planning, ensuring that the UAV autonomously avoids power lines and does not enter hazardous regions between adjacent towers.
We assume that the space near the tower is collision-free and trees, bushes, or any artificial infrastructure are not in close proximity to the tower.

Our method stores the positions of detected insulators in the world frame, preventing multiple detections of the same insulator.
This mechanism ensures that the UAV does not mistakenly interpret repeated observations of a single insulator as distinct objects.
In addition, the world-frame positioning is used to determine whether a newly detected insulator is located on the far side of the tower relative to the UAV.
Such insulators are not inspected, as the UAV focuses on navigating around and inspecting only the nearer side of the tower, and then inspects the farther side in the next step.
\section{Results}

The proposed online inspection algorithm was evaluated in terms of time efficiency, and its performance was demonstrated in a simulation and a real-world experiment. In addition, we perform a statistical evaluation of the detection precision and insulator localization error of the proposed methods.
Our approach was implemented and integrated into the multirotor UAV control and state-estimation framework developed by the Multi-robot Systems (MRS) Group \cite{baca2021mrs}.

\subsection{Simulation Results}\label{sec:simulation}


The YOLOv11n insulator detection framework was tested in order to verify detection precision. 
A detailed description of the dataset and the model training is provided in \refsec{sec:insulator_detection}. 
To measure the precision of the trained model, we chose the well-known mean Average Precision (mAP50–95) metric. 
The metric denotes average precision over multiple Intersection-over-Union (IoU) thresholds ranging from 0.50 to 0.95. 
It evaluates not only whether an object is detected, but also how accurately its bounding box aligns with the ground truth. 
The achieved mAP50–95 of 0.81 indicates successful training.



 
We compare the insulator localization errors of the proposed methods RANSAC, DBSCAN, DBSCAN + RANSAC, and DBSCAN + PCA, described in \refsec{sec:insulator_detection}.
The comparison is performed on measured data from two different towers A and B, depicted in \reffig{fig:towers}, and for different flight distances $w$ from the insulators 
to show the method's performance for different densities of incoming data.
Except for the density of points, parameter $w$ also influences the number of outliers in the detected bounding box as the probability of obtaining points representing reflections from the thin parts of the tower construction and power lines increases with decreasing sensing distance. 
Testing different tower types demonstrates that the proposed methods can localize insulators with different mechanical solutions for attaching them to the tower structure.
Testing data consists of 150 filtered point clouds.
Values of $w$ in tables were chosen with respect to the size of the tower and represent realistic values for real-world inspection. 
\reftab{tab:combined_error_A_transposed} and \reftab{tab:combined_error_B_transposed} show a comparison of the mean insulator localization errors for towers A and B, respectively.

\begin{table}[htbp]
    \centering
    \caption{Insulator localization error for a changing distance from insulator $w$ on tower A. The statistics are computed based on 150 measurements.}
    \begin{tabular}{|c||c|c|c|}
    \hline
    Method & $w = 8$ [m] & $w= 9$ [m] & $w= 10$ [m] \\
    \hline\hline
    RANSAC & 0.40 $\pm$ 0.20 & 0.40 $\pm$ 0.24 & 0.43 $\pm$ 0.30 \\ \hline
    DBSCAN + RANSAC & \textbf{0.35 $\pm$ 0.17} & \textbf{0.37 $\pm$ 0.23} & \textbf{0.38 $\pm$ 0.22} \\ \hline
    DBSCAN & 0.74 $\pm$ 0.57 & 0.66 $\pm$ 0.47 & 0.50 $\pm$ 0.33 \\ \hline
    DBSCAN + PCA & 0.79 $\pm$ 0.65 & 0.60 $\pm$ 0.57 & 0.40 $\pm$ 0.35 \\ \hline
    \end{tabular}
    \label{tab:combined_error_A_transposed}
    \vspace{-0.3cm}
\end{table}
\begin{table}[htbp]
    \centering
    \caption{Insulator localization error for a changing distance from insulator $w$ on tower B. The statistics are computed based on 150 measurements.}
    \begin{tabular}{|c||c|c|}
    \hline
    Method & $w  = 8 $ [m] & $w  = 9$ [m] \\
    \hline\hline
    RANSAC & 0.39 $\pm$ 0.24 & 0.43 $\pm$ 0.21 \\ \hline
    DBSCAN + RANSAC & 0.34 $\pm$ 0.20 & \textbf{0.36 $\pm$ 0.23} \\ \hline
    DBSCAN & \textbf{0.32 $\pm$ 0.21} & 0.41 $\pm$ 0.25 \\ \hline
    DBSCAN + PCA & 0.33 $\pm$ 0.21 & 0.41 $\pm$ 0.26 \\ \hline
    \end{tabular}
    \label{tab:combined_error_B_transposed}
\end{table}

The best performing method DBSCAN + RANSAC localizes insulators with an overall mean error of 0.36 $\pm$ 0.21 $\mathrm{m}$. This accuracy is sufficient for use as a positional reference for image acquisition and can serve as information for maintenance purposes. 
Compared to DBSCAN + PCA, the DBSCAN + RANSAC method achieves an overall lower insulator localization error by approximately 30 \%.
Smaller values of the parameter $w$ on tower A cause an increasing number of outliers, which the DBSCAN + PCA method fails to remove. 
The methods implementing RANSAC generally show a higher level of robustness. Specifically, DBSCAN + RANSAC shows increased precision by about 14 \% compared to RANSAC alone on tower B.
According to our observation, this advantage is given by the fact that the DBSCAN generally can better filter out points representing reflections from power lines near the insulator.

\vspace{-1em}
\begin{figure}[t]
  \centering
  \includegraphics[width=0.30\textwidth]{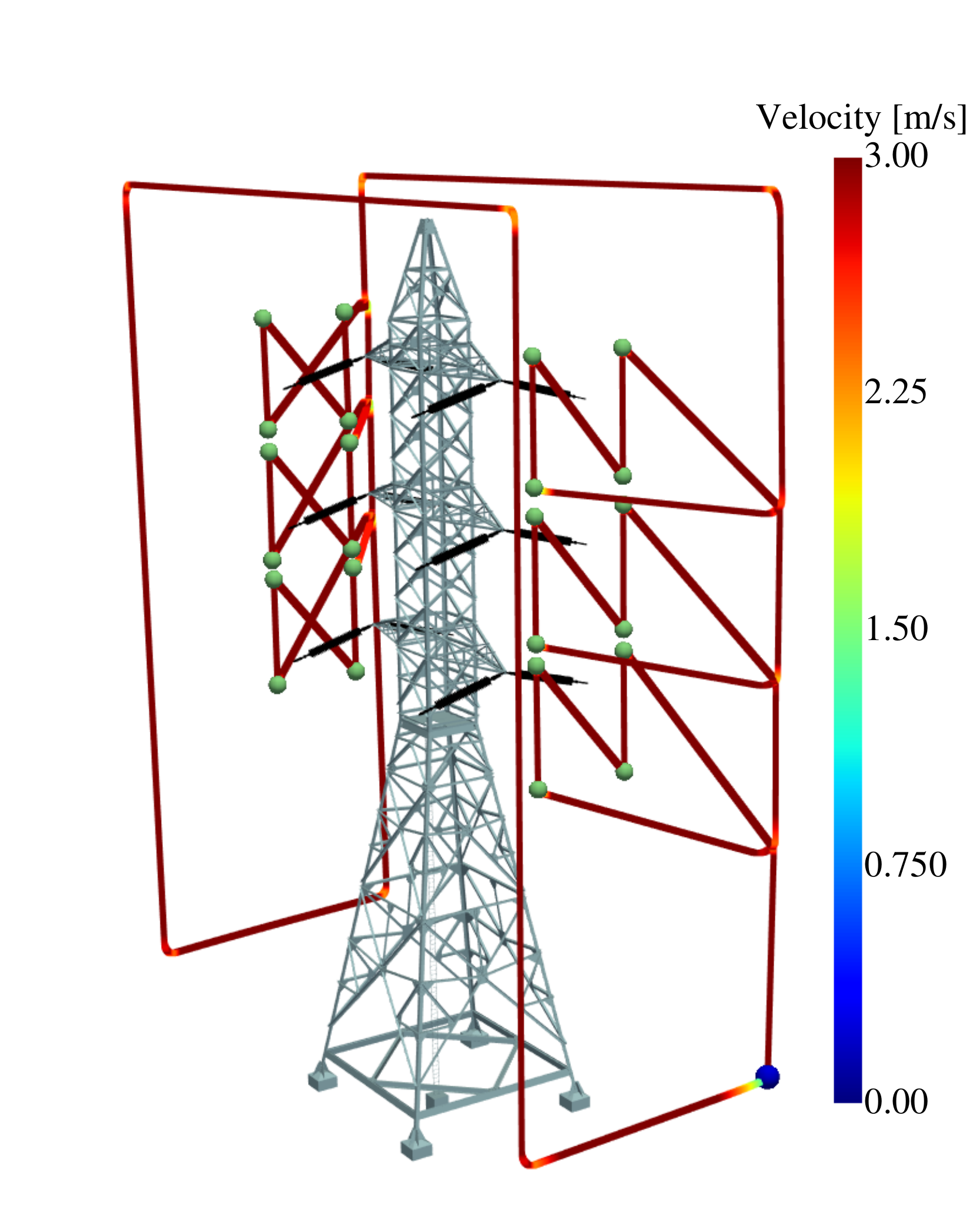}  
  \vspace{-0.5em}
  \caption{Trajectory of the UAV guided by our online inspection algorithm in Flight Forge simulator \cite{capek2025flightforge} computed by PMM planner \cite{teissing2024pmm}. Green points are the inspection waypoints suitable for obtaining detailed images of insulators. The starting position is depicted as a blue point.}
  \label{fig:simulation_trajectory}
  \vspace{-0.65cm}
\end{figure} 
\begin{figure}[ht]
  \centering
  \subfloat[Tower B] {
    \includegraphics[height=0.20\textwidth]{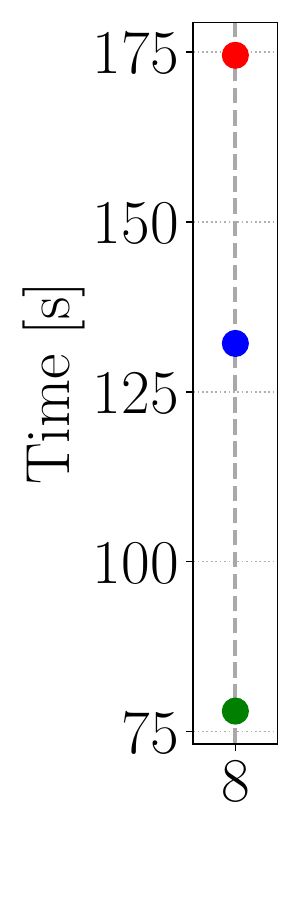}
    \label{fig:efficiency_tower_b}
  }
  \subfloat[Tower A] {
    \includegraphics[height=0.20\textwidth]{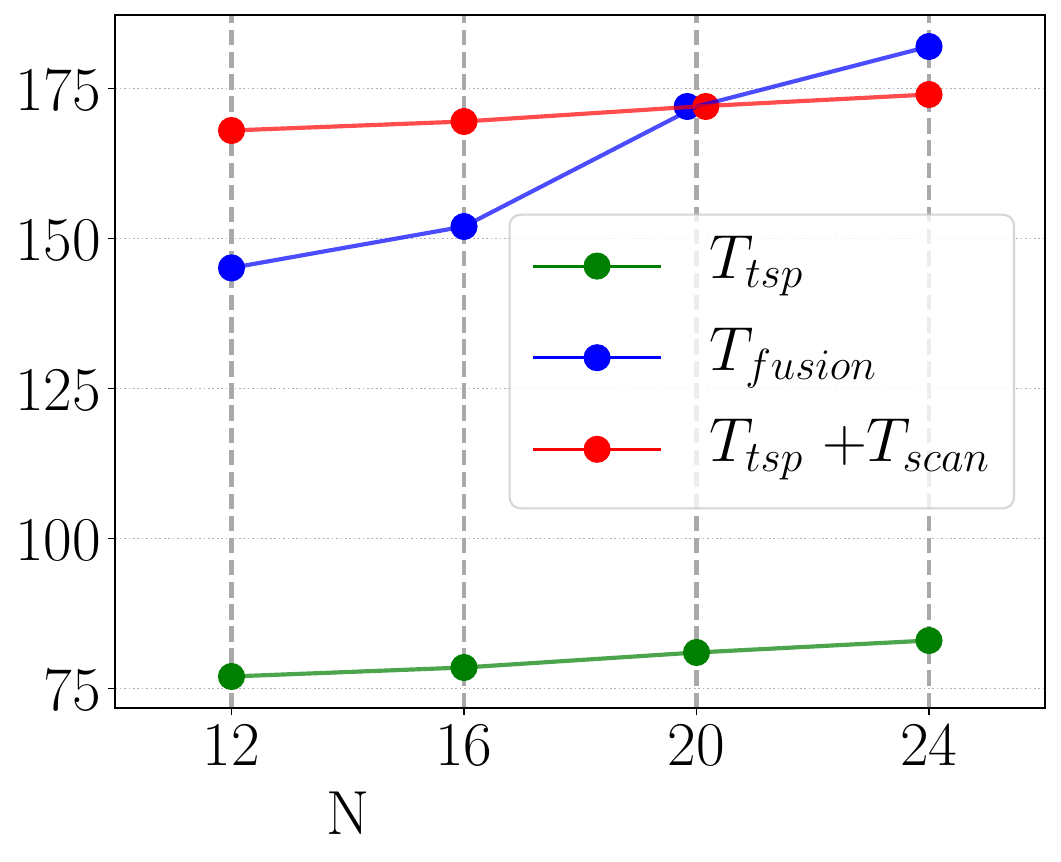}
    \label{fig:efficiency_tower_a}
  }
  \caption{Comparison of the duration $T_{fusion}$ achieved by our approach with
    the duration $T_{tsp}$ as optimal solution of TSP. The combined $T_{tsp}$ plus $T_{scan}$ of scanning the tower is also shown and included in the comparison. (a) depicts durations for inspection of tower A, (b) depicts durations for inspection of tower B. }
  \label{fig:efficiency_tower}
  \vspace{-0.45cm}
\end{figure}

Apart from the insulator localization precision, we further compare the duration of the inspection mission using the proposed online inspection algorithm in simulation with the duration of two-flight strategy with first scanning flight that creates map, followed by a time-optimal flight.
Specifically, the created map from scanning is used to localize the insulators and consequently generate the inspection waypoints. These are then passed to a Travelling Salesman Problem (TSP) solver to get the optimal sequence of visits, minimizing the total trajectory duration.
 To solve the TSP, the cost of edges was computed using the point mass model (PMM) time-optimal planner \cite{teissing2024pmm}. 
 To ensure collision-free trajectories between positions on the other side of the tower, additional waypoints above the tower were included.
 To guarantee a fair comparison, the PMM planner \cite{teissing2024pmm} was also used to compute individual trajectories for our single-flight approach, as depicted in \reffig{fig:simulation_trajectory}.
 The maximum velocity was set to 3 $\mathrm{ms^{-1}}$ and the maximum acceleration to 12 $\mathrm{ms^{-2}}$. We also compare the inspection duration for different numbers of inspection waypoints $N$.
 The comparison of durations is shown in \reffig{fig:efficiency_tower}. 
 The results demonstrate that the proposed method is considerably more efficient for a smaller number of inspection waypoints $N$ compared to the independent scanning flight followed by the optimal inspection flight.
 For example, our method saves 24 \% of time compared to the two-flight approach regarding $N=8$.  
The approach shows natural limitations for large values of $N$, as the UAV must repeatedly pause the exploration to visit new positions. 
However, even for $N=24$, the proposed single-flight strategy requires only 4.6 \% of additional time compared to the two-flight strategy. 

\subsection{Real-world Validation}

The proposed online inspection algorithm was implemented on a custom-designed UAV and validated through a real-world experiment conducted on a high-voltage transmission tower.
This experiment aimed to demonstrate the functionality of the system in real operating conditions. 
Measured data were also used to evaluate the precision of insulator detection and the accuracy of its localization.
The algorithm was deployed on a Jetson Orin Nano 8 GB high-performance computing unit equipped with a 1.5 GHz six-core Arm Cortex-A78AE CPU and a 625 MHz 1024-core NVIDIA GPU.
The UAV was equipped with a Livox Mid-360 LiDAR sensor, a Basler Daa1600-60uc camera, and a GoPro 12 mounted on a gimbal.
The Basler camera was used for insulator detection, while the GoPro was used for inspection images.
The UAV was equipped with Pixhawk 6C mini with PX4 low-level flight controller and Holybro H-RTK F9P GNSS Series RTK GPS receiver.

Dynamic constraints and inspection parameters used for the real-world experiment are shown in \reftab{tab:params_limits}. Maximum height $h_{max}$ and width $w_{max}$ of the tower were set by manual measurements. Constraints and parameters were set to secure a safe flight of the UAV near the critical infrastructure.

\begin{table}[h]
  \centering
  \caption{Dynamic constraints and inspection parameters \label{tab:params_limits}}
  \begin{tabular}{l c | l c | l c}
    \toprule
    \multicolumn{2}{c}{Horizontal constraints} & \multicolumn{2}{c}{Vertical constraints} & \multicolumn{2}{c}{Inspection parameters} \\
    \midrule
    $\mathbf{v}_{max}$ [$\mathrm{ms^{-1}}$] & 1 & $\mathbf{v}_{max}$ [$\mathrm{ms^{-1}}$] & 1 & $h_{max}$ \hfill [$\mathrm{m}$] & 25 \\
    $\mathbf{a}_{max}$ [$\mathrm{ms^{-2}}$] & 3 & $\mathbf{a}_{max}$ [$\mathrm{ms^{-2}}$] & 3 & $w_{max} $ \hfill [$\mathrm{m}$] & 10 \\
    \bottomrule
  \end{tabular}
  \vspace{-0.7em}
\end{table}

\begin{figure}[h]
\centering
\begin{minipage}{0.778\linewidth}
\begin{tikzpicture}
  \node[anchor=south west,inner sep=0] (a) at (0,0) {
    \includegraphics[width=\linewidth,
      trim={0cm 2cm 0cm 2cm}, clip]{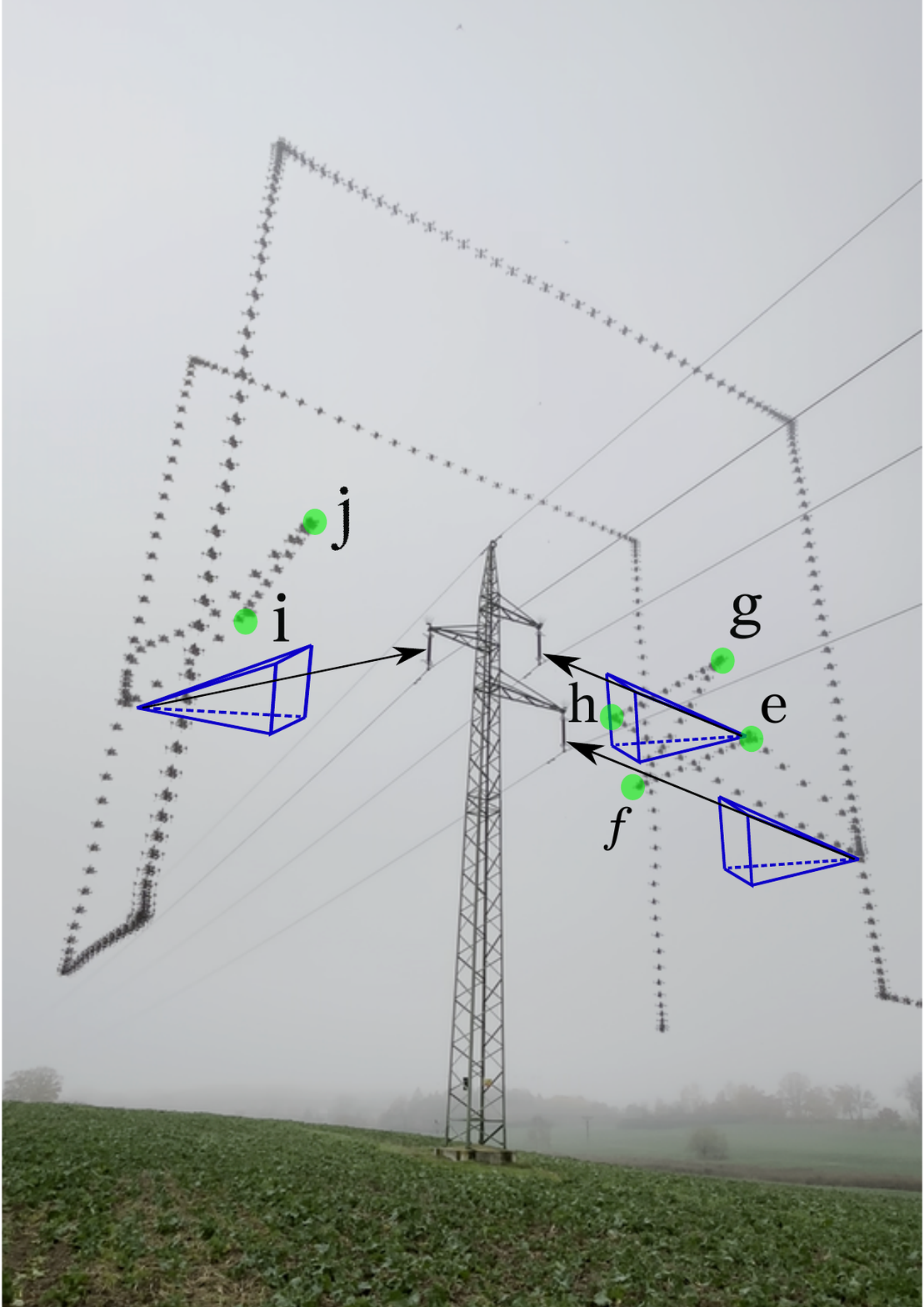}};
  \draw[white, very thick] (a.south west) rectangle (a.north east);
  \fill[white, opacity=0.8] (0.4, 0.4) circle (0.3cm);
  \draw (0.4,0.4) node {\small (a)};
\end{tikzpicture}
\end{minipage}
\hfill
\begin{minipage}{0.208\linewidth}

\begin{tikzpicture}
  \node[anchor=south west,inner sep=0] (a) at (0,0) {
    \includegraphics[width=\linewidth]{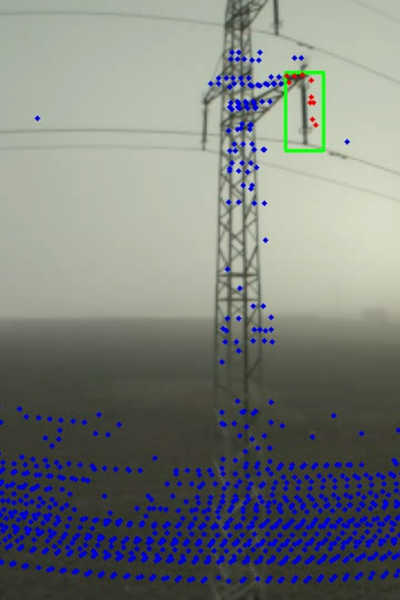}};
  \draw[white, very thick] (a.south west) rectangle (a.north east);
  \fill[white, opacity=0.8] (0.4, 0.4) circle (0.3cm);
  \draw (0.4,0.4) node {\small (b)};
\end{tikzpicture}

\vspace{0.01cm}

\begin{tikzpicture}
  \node[anchor=south west,inner sep=0] (a) at (0,0) {
    \includegraphics[width=\linewidth]{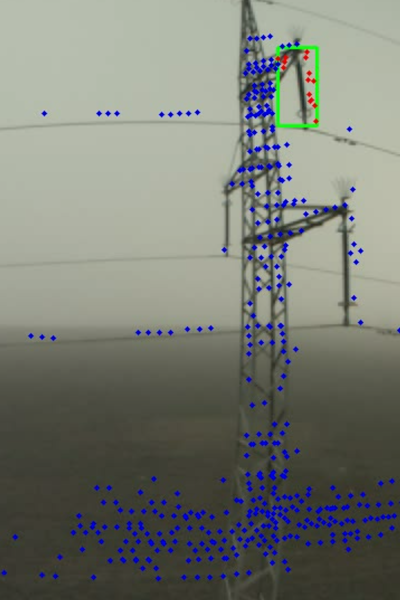}};
  \draw[white, very thick] (a.south west) rectangle (a.north east);
  \fill[white, opacity=0.8] (0.4, 0.4) circle (0.3cm);
  \draw (0.4,0.4) node {\small (c)};
\end{tikzpicture}

\vspace{0.01cm}

\begin{tikzpicture}
  \node[anchor=south west,inner sep=0] (a) at (0,0) {
    \includegraphics[width=\linewidth]{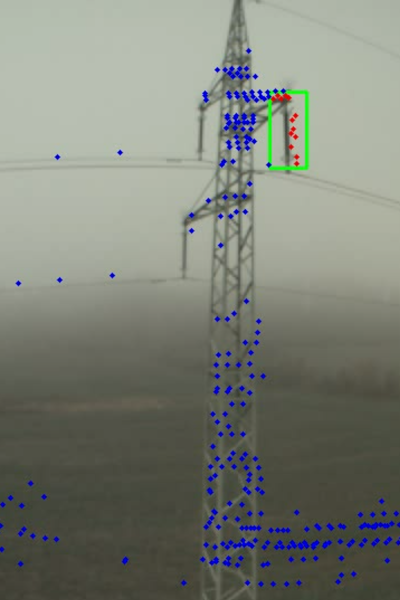}};
  \draw[white, very thick] (a.south west) rectangle (a.north east);
  \fill[white, opacity=0.8] (0.4, 0.4) circle (0.3cm);
  \draw (0.4,0.4) node {\small (d)};
\end{tikzpicture}
\end{minipage} 
\begin{minipage}{\linewidth}
\centering

\begin{tikzpicture}
  \node[anchor=south west,inner sep=0] (img) at (0,0)
  {\includegraphics[width=0.329\linewidth,
    trim=0 4cm 0 6cm, clip]
    {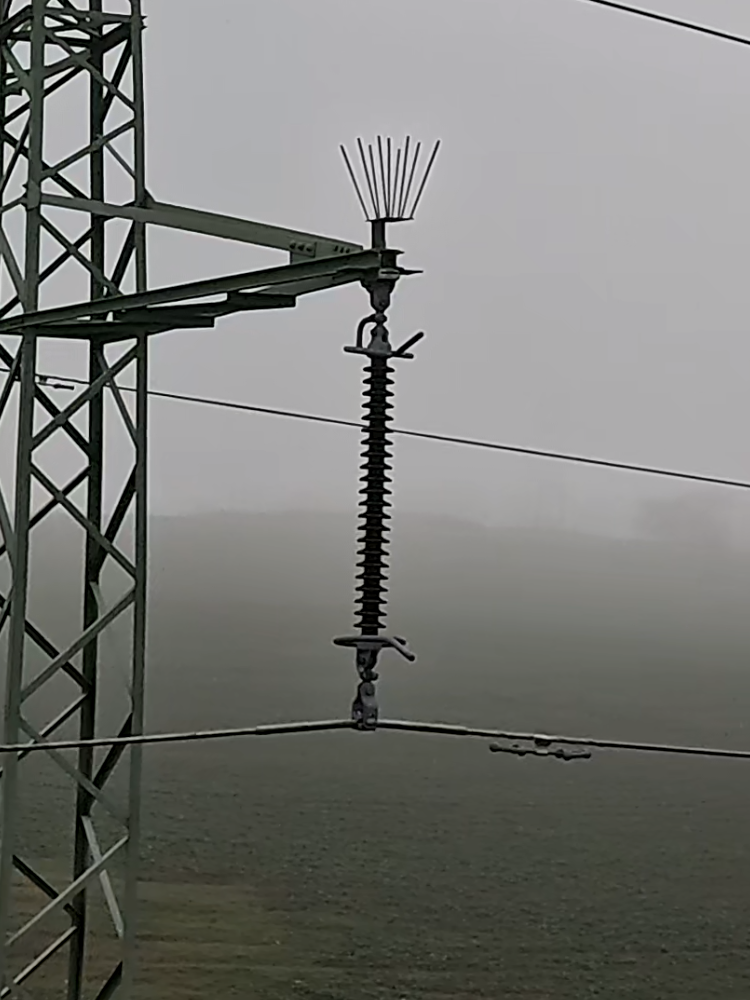}};
  \draw[white, very thick] (img.south west) rectangle (img.north east);
  \fill[white, opacity=0.8] (0.4,0.4) circle (0.3cm);
  \node at (0.4,0.4) {\small (e)};
\end{tikzpicture}%
\begin{tikzpicture}
  \node[anchor=south west,inner sep=0] (img) at (0,0)
  {\includegraphics[width=0.329\linewidth,
    trim=0 4cm 0 6cm, clip]
    {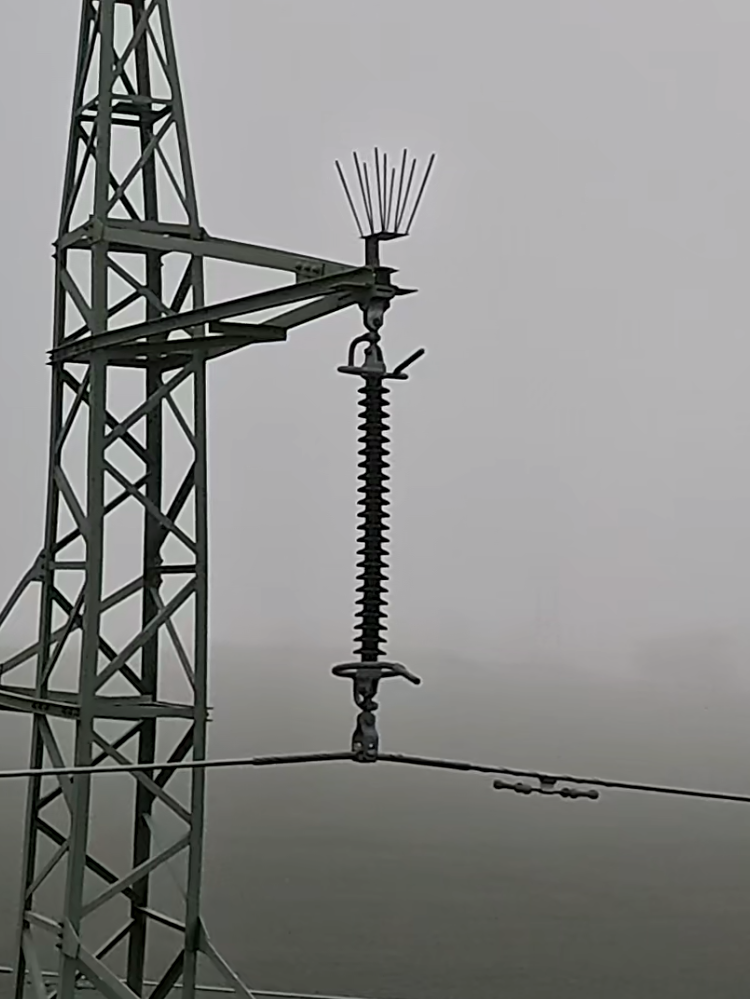}};
  \draw[white, very thick] (img.south west) rectangle (img.north east);
  \fill[white, opacity=0.8] (0.4,0.4) circle (0.3cm);
  \node at (0.4,0.4) {\small (g)};
\end{tikzpicture}%
\begin{tikzpicture}
  \node[anchor=south west,inner sep=0] (img) at (0,0)
  {\includegraphics[width=0.329\linewidth,
    trim=0 4cm 0 6cm, clip]
    {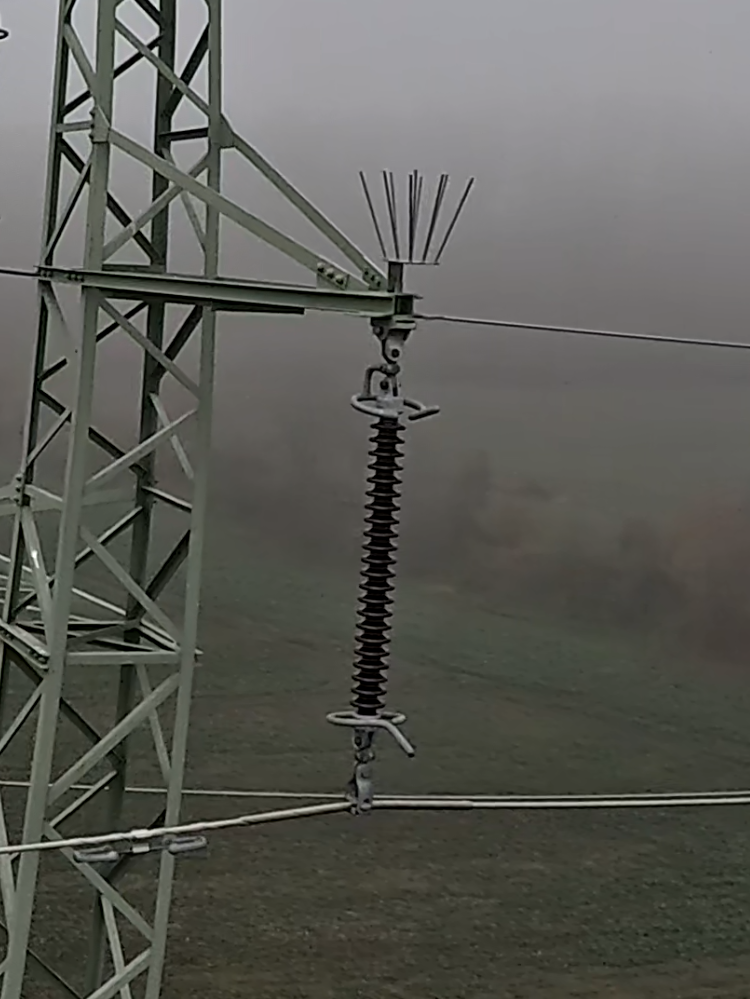}};
  \draw[white, very thick] (img.south west) rectangle (img.north east);
  \fill[white, opacity=0.8] (0.4,0.4) circle (0.3cm);
  \node at (0.4,0.4) {\small (i)};
\end{tikzpicture}%

\vspace{0.01cm}

\begin{tikzpicture}
  \node[anchor=south west,inner sep=0] (img) at (0,0)
  {\includegraphics[width=0.329\linewidth,
    trim=0 4cm 0 6cm, clip]
    {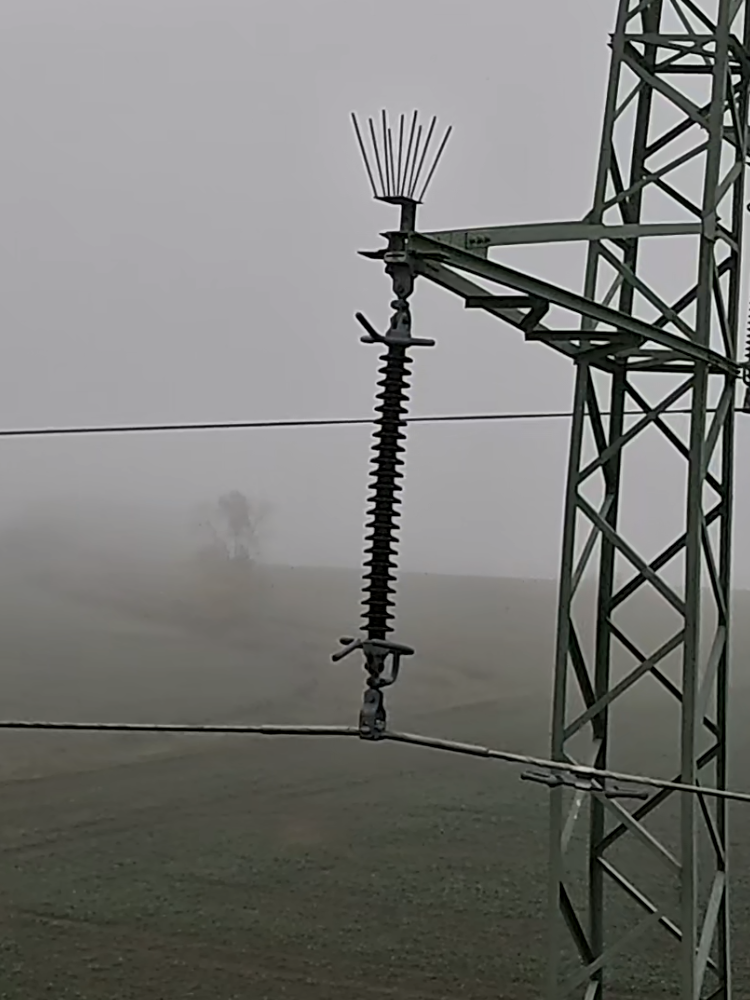}};
  \draw[white, very thick] (img.south west) rectangle (img.north east);
  \fill[white, opacity=0.8] (0.4,0.4) circle (0.3cm);
  \node at (0.4,0.4) {\small (f)};
\end{tikzpicture}%
\begin{tikzpicture}
  \node[anchor=south west,inner sep=0] (img) at (0,0)
  {\includegraphics[width=0.329\linewidth,
    trim=0 4cm 0 6cm, clip]
    {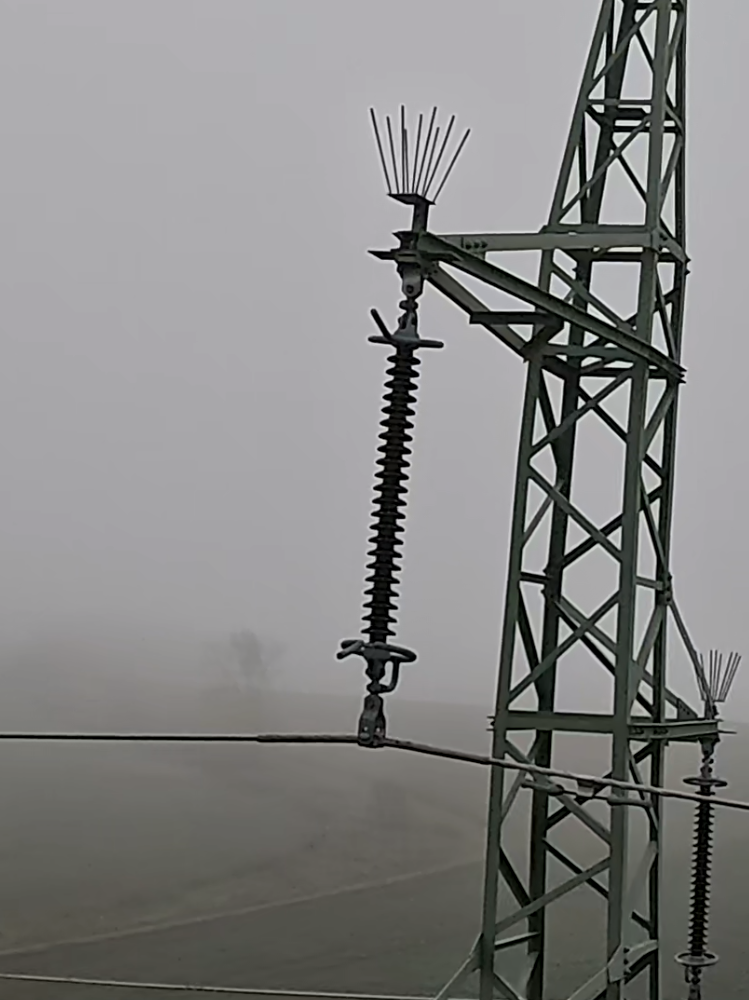}};
  \draw[white, very thick] (img.south west) rectangle (img.north east);
  \fill[white, opacity=0.8] (0.4,0.4) circle (0.3cm);
  \node at (0.4,0.4) {\small (h)};
\end{tikzpicture}%
\begin{tikzpicture}
  \node[anchor=south west,inner sep=0] (img) at (0,0)
  {\includegraphics[width=0.329\linewidth,
    trim=0 4cm 0 6cm, clip]
    {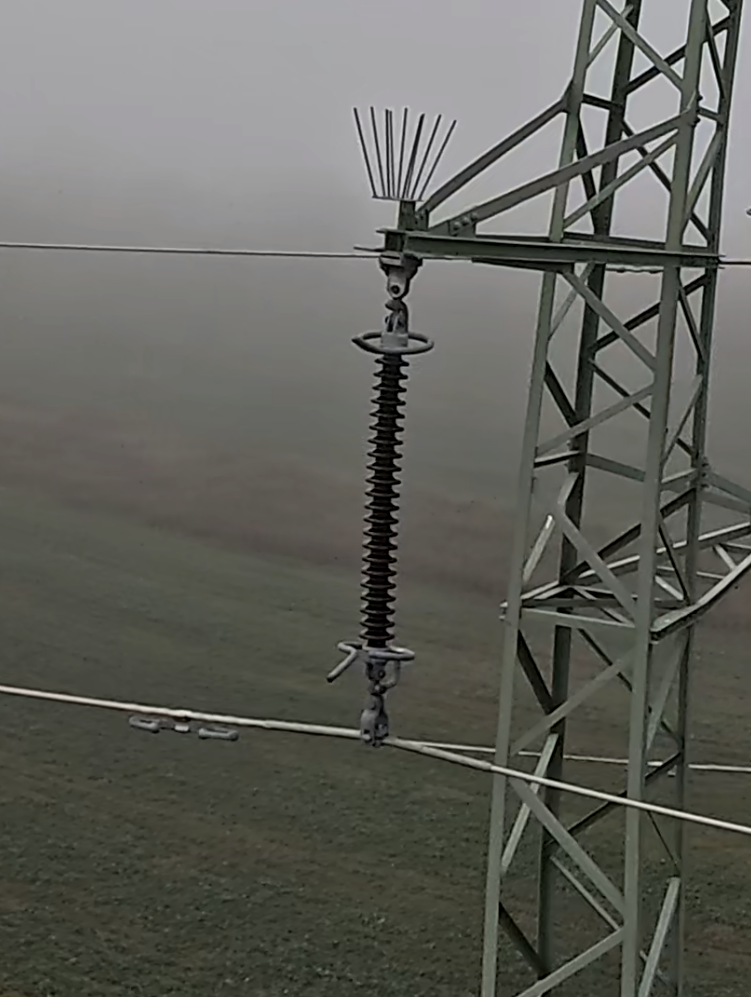}};
  \draw[white, very thick] (img.south west) rectangle (img.north east);
  \fill[white, opacity=0.8] (0.4,0.4) circle (0.3cm);
  \node at (0.4,0.4) {\small (j)};
\end{tikzpicture}%

\end{minipage}
\vspace{-0.5em}
\caption{The UAV guided by the online inspection algorithm with onboard sensor fusion in a real-world experiment. The blue FOV frustums illustrate the detection of each insulator in the picture (a). Each detection is also shown in the right column of images (b) - (d) together with the bounding box and the visualized filtered point cloud. The green markers in (a) represent the inspection waypoints used for image capturing and captured inspection images are shown in (e) - (j). } 

\label{fig:real_flight}
\vspace{-1em}
\end{figure}

The real-world YOLOv11n detection framework was verified independently to evaluate insulator detection precision.
We chose the same metric as for the simulation detection mAP50–95 described in \refsec{sec:simulation}.
The achieved mAP50–95 is 0.17.
This value shows a relative decrease of the metric of more than 79 \% compared to the simulation evaluation.
This lowered performance is caused mainly by harsh conditions of the test data, such as fogged camera lenses and non-ideal light conditions (see \reffig{fig:real_flight}), which were not sufficiently reflected in the diversity of the training dataset.

\begin{figure}[t]
  \centering
  \subfloat[plane $xy$] {
    \includegraphics[width=0.22\textwidth]{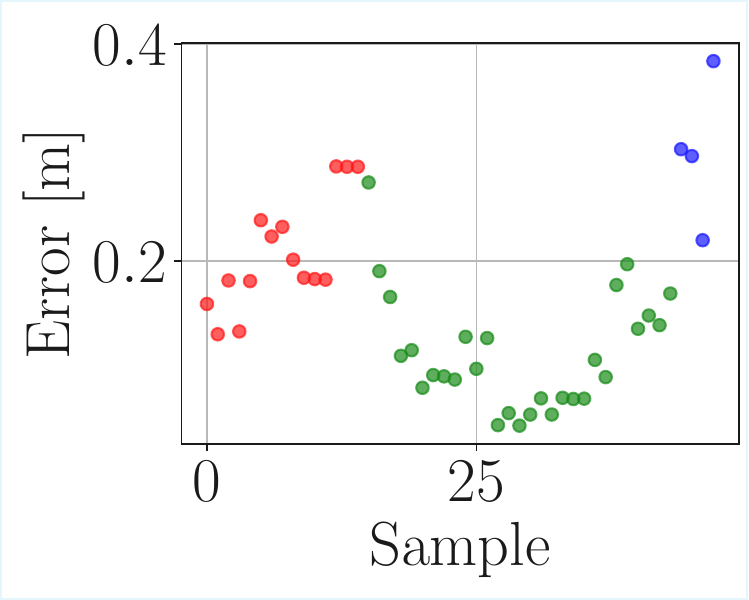}
    \label{fig:localization_error_xy}
  }
  \subfloat[altitude $z$] {
    \includegraphics[width=0.224\textwidth]{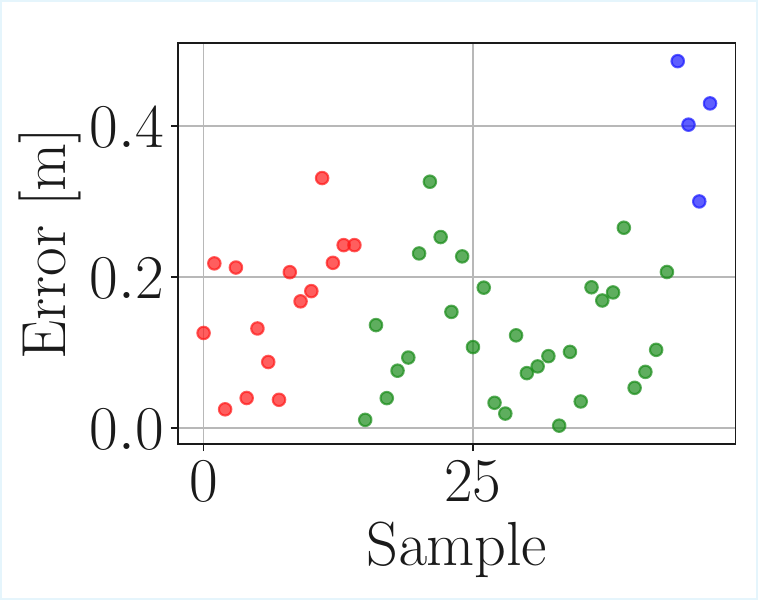}
    \label{fig:localization_error_z}
  }
  \caption{Real-world insulator localization errors for $xy$ plane in (a) and altitude $z$ in (b). Different colors correspond to different instances of an insulator.}
  \label{fig:localization_error_real}
  \vspace{-0.45cm}
\end{figure}

The performance of the online inspection algorithm, together with the best-performing localization method (DBSCAN + RANSAC), was tested onboard the UAV platform in a real-world high-voltage tower inspection scenario.
The experiment is depicted in \reffig{fig:real_flight}. 
The sensor fusion pipeline was evaluated against manually localized insulators inside a full LiDAR scan reconstructed using the iterative closest point (ICP) algorithm. 
The scan was aligned with the measured ground-truth tower position.
The mean localization error is $0.16 \pm 0.08$ m in the $xy$ plane and $0.16 \pm 0.11$ m in altitude $z$.
In terms of standard deviation error in the $xy$ plane, our method achieves one order of magnitude higher accuracy than the approach based on a binocular camera \cite{rs13020230}.
Insulator localization errors in the $xy$ plane and altitude $z$ during flight are shown in \reffig{fig:localization_error_real}. 
Sufficient accuracy in insulator localization is also confirmed by the captured (and additionally cropped) images of the insulators shown in \reffig{fig:real_flight}.
\vspace{-0.2cm}

\section{Conclusion}

    We introduced the online inspection algorithm that enables inspection of an unmapped tower within a single flight without information about the exact number or positions of insulators. Our algorithm relies on the camera-LiDAR fusion pipeline for onboard insulator detection and localization. 
YOLOv11n was used for insulator detection. For insulator localization, we proposed and compared different methods based on DBSCAN, RANSAC, or PCA algorithms.
Our approach was verified in both simulation and real-world experiments.
In simulation, we proved that our single-flight inspection strategy can save up to 24 \% of total inspection time, compared to the two-flight approach of mapping the tower followed by visiting the detected inspection waypoints in the optimal way.
The proposed best-performing DBSCAN+RANSAC method achieves a real-world localization error of $0.16 \pm 0.08$ m in the $xy$ plane and $0.16 \pm 0.11$ m in altitude $z$, showing that our approach is capable of accurate insulator localization for the acquisition of inspection images.\vspace{-0.1cm}

\bibliographystyle{IEEEtran}
\bibliography{references}


\vspace{12pt}

\end{document}